\documentclass{article}

 \usepackage[preprint]{neurips_2025}

\usepackage[utf8]{inputenc} 
\usepackage[T1]{fontenc}    
\usepackage{hyperref}       
\usepackage{url}            
\usepackage{booktabs}       
\usepackage{amsfonts}       
\usepackage{nicefrac}       
\usepackage{microtype}      
\usepackage{xcolor}         
\usepackage{graphicx}
\usepackage{amsmath}
\usepackage{subcaption}
\usepackage{multirow}
\usepackage{tabularx}
\title{Towards Fully Automated Molecular Simulations: Multi-Agent Framework for Simulation Setup and Force Field Extraction}
\workshoptitle{AI4Mat: AI for Accelerated Materials Design}

\author{%
  Marko Petkovi\'{c}\\\
  Applied Physics and Science Education\\
  Eindhoven University of Technology\\
  \And
  Vlado Menkovski \\
  Mathematics and Computer Science \\
  Eindhoven University of Technology \\
  \AND
  Sof\'{i}a Calero\\\
  Applied Physics and Science Education\\
  Eindhoven University of Technology\\
  \texttt{\{m.petkovic1, v.menkovski, s.calero\}@tue.nl} \\
}

\begin{document}

\maketitle

\begin{abstract} 
Automated characterization of porous materials has the potential to accelerate materials discovery, but it remains limited by the complexity of simulation setup and force field selection. We propose a multi-agent framework in which LLM-based agents can autonomously understand a characterization task, plan appropriate simulations, assemble relevant force fields, execute them and interpret their results to guide subsequent steps. As a first step toward this vision, we present a multi-agent system for literature-informed force field extraction and automated RASPA simulation setup. Initial evaluations demonstrate high correctness and reproducibility, highlighting this approach’s potential to enable fully autonomous, scalable materials characterization.
\end{abstract}

\section{Introduction}
Porous materials, such as metal–organic frameworks (MOFs) and zeolites, are central to applications in energy storage, gas separation, catalysis, and carbon capture \cite{corma1997microporous,furukawa2013chemistry,li2009selective}. The design of novel porous materials with improved performance can have significant societal and environmental benefits. Molecular simulations play an important role in this process, as they provide molecular-level insights into adsorption and diffusion that are often inaccessible to experiments \cite{coudert2015responsive,duren2007calculating,smit2008molecular}. By revealing how structural features influence adsorption and transport, simulations can identify promising candidates and guide experimental efforts, thereby accelerating the discovery of porous materials \cite{colon2014high,wilmer2012large}.

The adoption of molecular simulations for porous materials is limited by several technical challenges. Outcomes are highly sensitive to the choice of force fields and charge assignment methods, with studies showing that different protocols can result in markedly different adsorption predictions and even alter material rankings \cite{cleeton2023process}. Maintaining reproducibility is also difficult, as variations in workflows, such as charge schemes, cutoff definitions, or structure curation, may hinder cross-study comparability \cite{ongari2020too}. Moreover, the diversity of force fields and their associated parameters requires deep expertise to select and apply them correctly \cite{farmahini2021performance}. These challenges create a barrier for wider use of simulations as a routine characterization tool. 

Automated characterization of porous materials presents a multifaceted challenge, primarily due to the necessity for intelligent planning. Agents must determine the appropriate simulations to conduct, adapt workflows based on interim results, select or design accurate force fields, and generalize across diverse topologies, adsorbates, and conditions. Recent advances in large language models (LLMs) have demonstrated their capability to address these complexities \cite{boiko2023autonomous,m2024augmenting,ruan2024automatic}. LLMs have been shown to effectively generate structured workflows, interpret complex scientific tasks and synthesize executable simulation code, thereby facilitating the automation of intricate scientific processes \cite{xu2024llm4workflow,zou2025agente,yoshikawa2023large,darvish2025organa}. Their proficiency in integrating heterogeneous knowledge and performing flexible reasoning and multi-step planning makes them well-suited for orchestrating the adaptive workflows required in materials characterization \cite{bubeck2023sparks,huang2024understanding}. Furthermore, LLMs have been utilized to automate research workflows, including experimental protocol planning and scientific code generation, underscoring their potential in scientific automation \cite{cao2025large,luo2025llm4sr,ren2025towards,wei2025plangenllms}.

\begin{figure}[h]
    \centering
    \includegraphics[width=.65\linewidth]{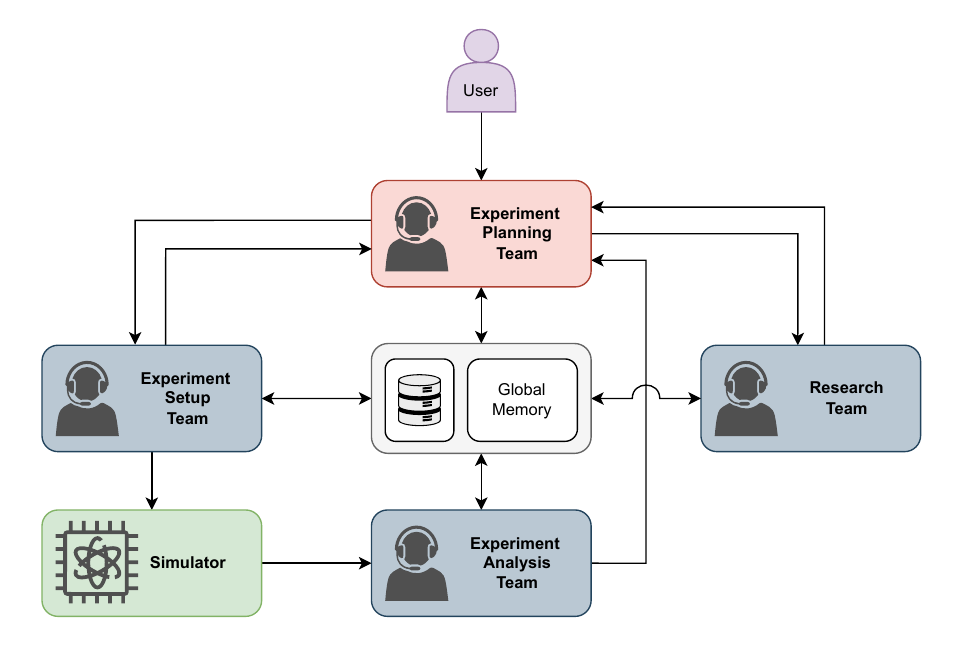}
    \caption{Vision of an autonomous multi-agent system for molecular simulations. The Experiment Planning Team coordinates research, setup, and analysis teams through a shared global memory,enabling iterative workflows in which user queries result in literature-derived force fields, RASPA simulations, and analysis of results.}
    \label{fig:vision}
\end{figure}

Our goal is to develop a system that, given an experimental measurement or user request, can autonomously plan and execute the necessary steps to identify or design suitable force fields and then perform characterization simulations (Figure \ref{fig:vision}). By automating both the selection of physically accurate force fields and the execution of complex simulation workflows, such a system bridges experimental observations with predictive, simulation-driven insights. This approach not only reduces the expertise and time required to perform high-quality simulations but also enables routine, reproducible characterization of porous materials, accelerating materials discovery and guiding experiments with a level of efficiency and reliability that would be difficult to achieve manually.

As a step toward this vision, we present two key agent-based components that form the foundation of the system. First, experiment setup agents that can automatically generate RASPA \cite{dubbeldam2016raspa} input files for increasingly complex characterization tasks, including adsorption isotherms, heats of adsorption, and multi-adsorbate or multi-structure scenarios. Second, force field retrieval agents that autonomously extract relevant parameters from the literature and convert them into a format ready for RASPA simulations. We evaluate the correctness and reliability of these components and discuss how they provide a scalable framework for fully automated characterization workflows in porous materials.

\section{Methods}

\begin{figure}[h!]
    \centering
    \includegraphics[width=.85\linewidth]{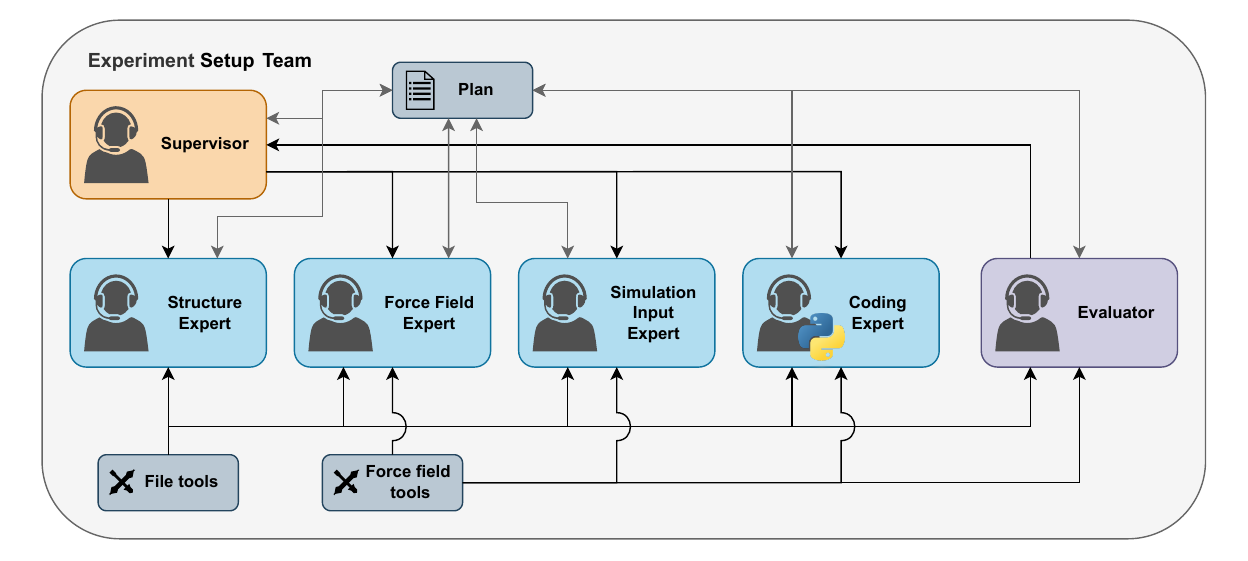}
    \caption{Experiment Setup Team overview. }
    \label{fig:simteam}
\end{figure}

Our system is organized into two specialized teams of LLM-based agents, using the ReAct framework \cite{yao2023react}, that collaboratively perform autonomous materials characterization. The experiment setup team (Figure \ref{fig:simteam}) is coordinated by a supervisor agent that processes user or experimental requests and sets up the corresponding simulations. The supervisor delegates responsibilities to specialized agents: a structure agent identifies and prepares relevant structure copies, a force field agent combines and formats appropriate force field files, a simulation input agent generates templated RASPA input files, and a code generation agent \cite{wang2024executable} automates file operations and template creation. After each agent completes its task, an evaluator agent \cite{zhuge2024agent} inspects the generated outputs and provides feedback to ensure correctness and consistency. Agents have access to general tools for file manipulation, while relevant agents also use functions to extract information from files without opening them directly. In addition, they share a global memory, which each agent can update with structured reports of task execution and outputs. Furthermore, agents have access to example input files from the RASPA manual, as well as a library containing multiple force fields to base their simulations on.

\begin{figure}[h!]
    \centering
    \includegraphics[width=.63\linewidth]{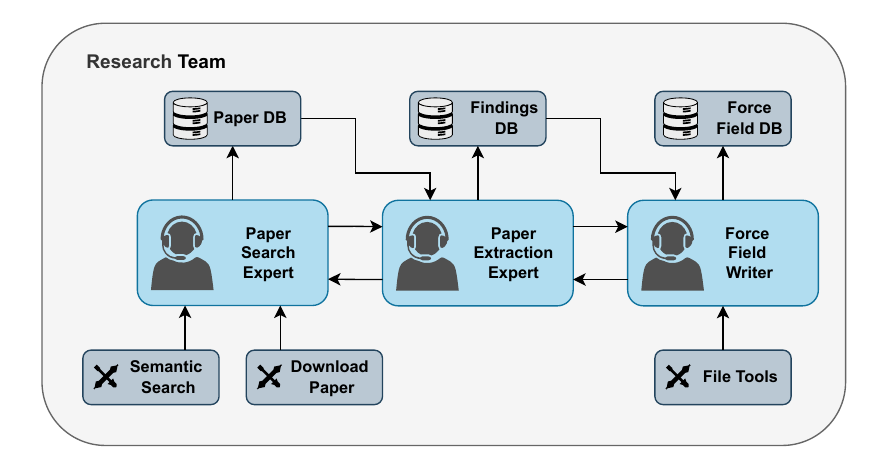}
    \caption{Research Team overview.}
    \label{fig:paperteam}
\end{figure}

The paper research team (Figure \ref{fig:paperteam}) extracts simulation-relevant knowledge from the literature. A paper search agent retrieves relevant publications, which are passed to an extraction agent that reads the papers and summarizes key findings related to force field parameters. A force field writer agent then converts these findings into simulation-ready force field files, and has access to dummy force field files to base the force fields on. Agents communicate iteratively, requesting additional information when needed (e.g., if a force field extends results from a different publication), ensuring accuracy and completeness of the extracted knowledge.

The agentic framework is implemented using LangChain and LangGraph. The paper search agent uses Semantic Scholar \cite{ammar2018construction} for queries. The code for the experiments can be found in this \href{https://anonymous.4open.science/r/sim_agent_anon-F7EC/}{anonymized repository}, while more details on the agentic system can be found in Appendix \ref{a:system}.

\section{Experiments}
To evaluate our simulation setup system, we designed a series of zeolite adsorption tasks with increasing complexity. These include adsorption isotherms and heats of adsorption (HOA) for single (CH$_4$) and multiple adsorbates (CH$_4$, CO$_2$, CO), as well as simulations spanning single and multiple zeolite structures. Each task is deliberately constructed so that agents must integrate input parameters from multiple sources, combining adsorbate properties and force field parameters from separate files, requiring correct assembly and templating. The system has access to several commonly used force fields for molecules in zeolites \cite{calero2004understanding,garcia2009transferable,martin2015transferable,romero2023adsorption}, which agents retrieve or combine as needed. Each simulation setup is performed five times, and the resulting files are evaluated both manually and by executing them in RASPA to verify correctness and reproducibility.

To evaluate the paper research team, we select a set of publications and process each five times to assess reproducibility. The agents extract force field parameters from the papers, and we verify whether all relevant parameters are correctly identified and formatted for use in simulations.

\begin{table}[h!]
\centering
\begin{minipage}[t]{0.45\textwidth}
\centering
\caption{Experiment Setup Team results. For each task, the number of zeolite structures and adsorbates, as well as the success and execution rates, are shown.}
\begin{tabular}{lcccc}
\toprule
\textbf{Task} & \textbf{Str.} & \textbf{Ads.} & \textbf{Succ.} & \textbf{Exec.} \\
\midrule
\multirow{2}{*}{Isotherm} & \multirow{2}{*}{1}   & 1 & 100\% & 100\% \\
                          &                       & 3 & 100\% & 100\% \\
\midrule
\multirow{2}{*}{HOA}      & \multirow{2}{*}{500} & 1 & 80\% & 100\% \\
                          &                       & 3 & 80\% & 80\%  \\
\midrule
\multirow{2}{*}{Isotherm} & \multirow{2}{*}{500} & 1 & 100\% & 100\% \\
                          &                       & 3 & 80\% & 100\%  \\
\bottomrule
\end{tabular}
\label{tab:simsetup}
\end{minipage}
\hfill
\begin{minipage}[t]{0.52\textwidth}
\centering
\caption{Research Team results. For each force field, the average number of missed parameters, wrong parameters, and the Intersection over Union (IoU) between extracted and correct parameters are shown.}
\begin{tabular}{lccc}
\toprule
\textbf{Force fields} & \textbf{Missed} & \textbf{Wrong} & \textbf{IoU} \\
\midrule
CO$_2$ \cite{garcia2009transferable} & 0 & 0 & 1.00 \\
TraPPE-zeo \cite{bai2013trappe}     & 0 & 0.6 & 0.90 \\
CO$_2$ N$_2$ O$_2$ Ar \cite{vujic2016transferable} & 0 & 1.4 & 0.96 \\
EPM2 (CO$_2$) \cite{harris1995carbon} & 0 & 3 & 0.67 \\
CO$_2$ \cite{fang2013first}     & 0 & 1 & 0.96 \\
C$_\text{x}$H$_\text{x}$ CO$_2$ N$_2$ O$_2$ \cite{findley2021transferable} & 0 & 0 & 1.00 \\
\bottomrule
\end{tabular}
\label{tab:paperteam}
\end{minipage}
\end{table}

Overall, the system performed well in the experiment setup task (Table \ref{tab:simsetup}), with a high rate of successful and executable simulations. We define the success rate as the proportion of simulations correctly configured for their intended task, and the execution rate as the proportion that run without errors. The few failures we observed came from specific issues: in the isotherm case with three adsorbates, the setup incorrectly generated a combined mixture simulation instead of three individual simulations; the Widom insertion method was not configured correctly for single-adsorbate HOA; and for three adsorbates, the framework file was not copied. We also observed certain decisions, such as selecting a non-standard cutoff (24 \AA \text{ }  instead of the typical 12 \AA) or enforcing minimum unit cells in each direction, that were unusual but still led to correct and runnable simulations. Additional details on the simulation setups, including deviations and encountered issues, are provided in Appendix \ref{a:runs}.

Table \ref{tab:paperteam} shows the performance of the Research Team in extracting force field parameters. Across all tested cases, no parameters were entirely missed, indicating high recall. Most errors arose from incorrect numerical values being assigned to otherwise correct terms. In the case of TraPPE-zeo, the model introduced a non-existent interaction term. For EPM2, all numbers were extracted correctly but assigned to the wrong interactions due to the unconventional table layout, which reduced the Intersection over Union (IoU) to 0.67. For the other force fields, the IoU remained high ($\geq 0.9$), showing that the system is generally reliable when parameter tables follow standard formats.

In addition to the independent experiments, we conducted a coordinated experiment in which the two teams were overseen by a supervisor. They were tasked with extracting the force field from \cite{garcia2009transferable} and setting up an adsorption isotherm simulation for a structure using the extracted force field. The system successfully completed this workflow. An overview of the execution is in Appendix \ref{a:c}.

\section{Discussion}
Our results demonstrate that autonomous agents can reliably set up molecular simulations and extract force field parameters from the literature, with only a small number of task-specific errors. These outcomes suggest that multi-agent systems provide a promising foundation for automated characterization of porous materials.

Looking forward, the most significant opportunity lies in equipping agent-based systems with structured, persistent memory representations. At present, knowledge is encoded procedurally in prompts and tools, limiting the ability of agents to adapt across tasks or to refine strategies based on past experience. Incorporating semantic memory (e.g., domain-specific heuristics that experts apply when setting up simulations) and episodic memory could enable more consistent and generalizable behavior \cite{sumers2023cognitive,huang2024understanding,zhuge2024agent}. This would allow agents to construct increasingly sophisticated workflows for materials characterization, bridging experimental observations with predictive modeling in a continuous loop. Ultimately, such systems could serve as the foundation for self-driving laboratories \cite{tom2024self}, where simulation and experimentation workflows are autonomously integrated in real time to accelerate the discovery of high-performance porous materials.

\newpage

\bibliographystyle{plain}
\bibliography{biblio}

\newpage
\appendix

\section{System Overview}\label{a:system}

\renewcommand{\thetable}{S\arabic{table}}  
\renewcommand{\thefigure}{S\arabic{figure}} 
\setcounter{figure}{0}
\setcounter{table}{0}

In Table \ref{tab:agentmodels}, a description of each agent, along with the LLM they are based on, can be found. Empirically, we found that most models performed significantly better when using gpt-5, compared to older models (gpt-4o, gpt-4.1).

\begin{table}[h!]
    \centering
    \caption{Overview of agents used in this work and models}
    \begin{tabularx}{\linewidth}{lXc}
        \toprule
        Agent Name & Description & LLM Model \\
        \midrule
        Supervisor & Responsible for understanding the user request, and developing a plan of actions needed to be performed to set up a simulation. Delegates to the various agents according to its plan. & gpt-5 \\\\
        Structure Expert & Finds the appropriate (placeholder) structure, and places a copy in the simulation folder. & gpt-5-mini \\\\
        Force Field Expert & Given a request, decides what the appropriate force fields are. If necessary, it combines them, and writes new force field files (e.g., \texttt{force\_field.def}, \texttt{pseudo\_atoms.def})  & gpt-5 \\\\
        Simulation Input Expert & Creates a \texttt{simulation.input} file, depending on the requirements of the simulation. Needs to decide which keywords are appropriate, and where to fill in numbers and to template. & gpt-5 \\\\
        Coding Expert & Writes code to replicate the template folder for each necessary run. Needs to understand how the simulation template is set up, and which fields need to be filled in and how. & gpt-5 \\\\
        Evaluator & Evaluates the task performance of each agent by inspecting files created during their execution and flags any potential mistakes it finds. & gpt-5 \\\\
        \midrule
        Paper Search Agent & Uses Semantic Scholar search to find appropriate research papers, and downloads them. & gpt-5-mini \\\\
        Paper Extraction Agent & Reads downloaded papers and extracts any relevant information from them (e.g., molecule definitions, interaction parameters). & gpt-5-mini \\\\
        Force Field Writer & Reads the findings produced by the paper extraction agent and transforms them into RASPA force field files.& gpt-5 \\\\
        \bottomrule
    \end{tabularx}
    \label{tab:agentmodels}
\end{table}

\begin{table}[h!]
    \centering
    \caption{Overview of tools available to various agents.}
    \begin{tabularx}{\linewidth}{lX}
        \toprule
        Tool Name & Description \\
        \midrule
        list\_all\_example\_simulation\_inputs & Gives the names and description of example input files from the RASPA manual \\\\
        
        read\_atoms\_in\_file & Returns the set of atoms present in a \texttt{framework.cif} or \texttt{molecule.def} file. \\\\
        
        count\_atom\_type\_in\_cif & Counts how often a given atom type occurs in a CIF file. \\\\
        
        get\_unit\_cell\_size & Returns the lattice parameters of the unit cell defined in a CIF file. \\\\
        
        get\_all\_force\_field\_descriptions & Lists the available force fields and their metadata. \\\\
        
        get\_atoms\_in\_ff\_file & Lists the set of atoms for which parameters are defined in a force field file (\texttt{force\_field.def},\texttt{ force\_field\_mixing\_rules.def}, \texttt{pseudo\_atoms.def}). \\\\

        semantic\_scholar\_search & Performs a search query using the Semantic Scholar. \\\\

        download\_paper & Downloads a paper given a DOI. \\\\

        read\_paper\_headers &  Lists the section headers of a paper after it has been parsed. \\\\

        read\_paper\_section & Returns the content of the section of a paper. \\\\
        
        \bottomrule
    \end{tabularx}
    \label{tab:agenttools}
\end{table}

An overview of the tools can be found in Table \ref{tab:agenttools}. In addition to the tools enumerated here, agents have access to various tools for reading, writing, and copying files.

\newpage
\section{Run Details}\label{a:runs}
In Table \ref{tab:rus}, a more detailed explanation can be found on issues and mistakes during the simulation setup process. Overall, most mistakes are minor and could be fixed in a more robust, future version of the system.

\begin{table}[h!]
    \centering
    
    \caption{Notes on simulation setups with mistakes/intricacies}
    \begin{tabularx}{\linewidth}{lccX}
         \toprule
        Task & Strucutre & Adsorbate & Notes \\
        \midrule
         Isotherm & 1 & 3 & In multiple runs, all adsorbate files were copied into each simulation folder. Only the correct adsorbate was used in \texttt{simulation.input}, resulting in correct simulations. \\\\
         HOA & 500 & 1 & In one of the runs, all possible moves were defined (with 0 probability, except widom insertions) for the adsorbate. In another run, no moves were specified for the adsorbate, resulting in an incorrect simulation where the adsorbate doesn't move. \\\\

        HOA & 500 & 3 & In one of the runs, the structure CIF files were not copied, resulting in failed simulations. In multiple runs, redundant/unused files were left as part of the force field. \\\\

        Isotherm & 500 & 1 & In one of the runs, a minimum amount of unit cells was enforced, resulting in many structures with more unit cells than necessary, resulting in slower simulations.\\\\

        Isotherm & 500 & 3 & Rather than generating separate simulations for all three adsorbates, a mixture isotherm was calculated in one of the simulations. While this is a valid simulation, this was not requested in the task. In another run, a cutoff of 24 \AA\text{} was selected, resulting in twice as many unit cells.\\

         \bottomrule
    \end{tabularx}
    \label{tab:rus}
\end{table}

\newpage
\section{Combined System Run}\label{a:c}

\begin{figure}[h!]
    \centering
    \includegraphics[width=0.9\linewidth]{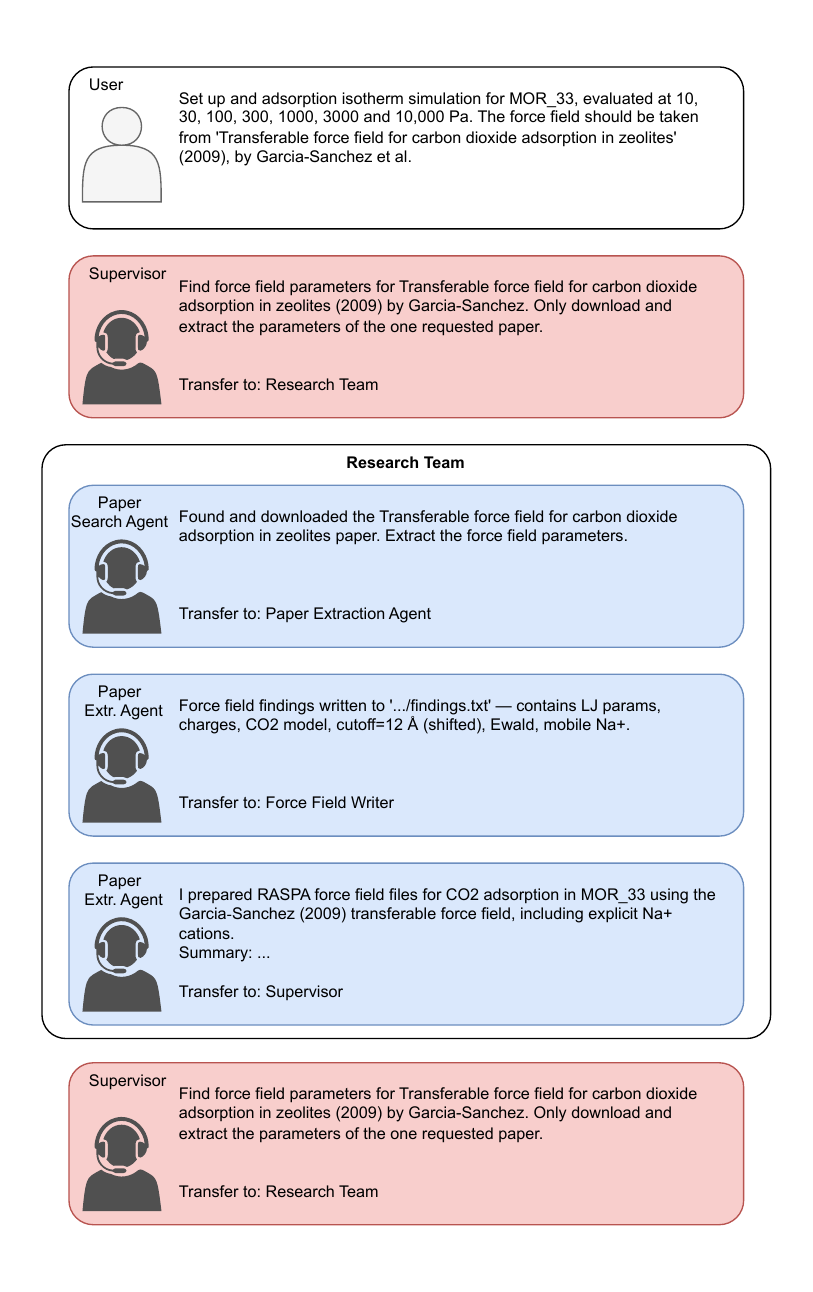}
    \caption{First part of the trace of the full system run.}
    \label{fig:placeholder1}
\end{figure}

\begin{figure}[h!]
    \centering
    \includegraphics[width=0.9\linewidth]{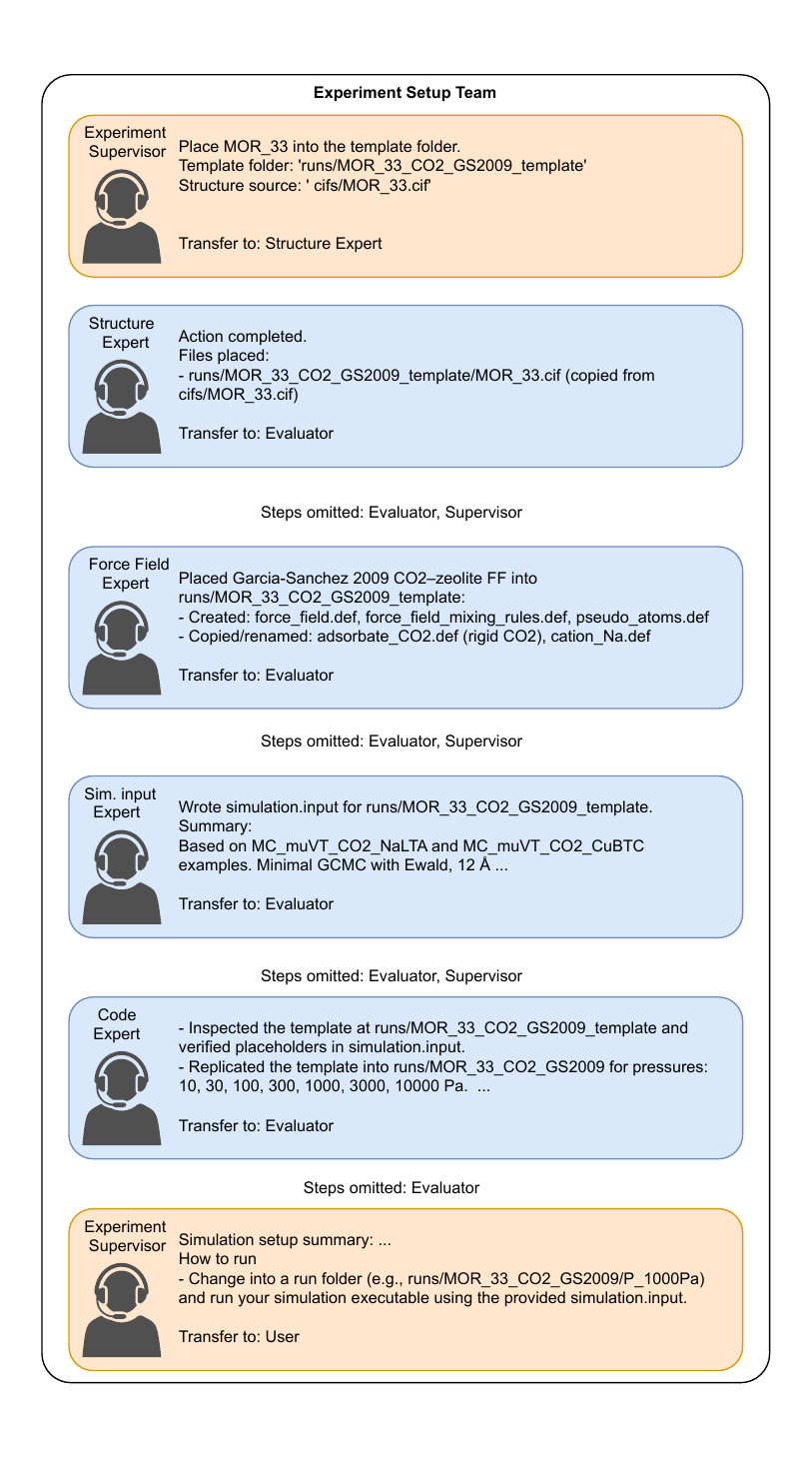}
    \caption{Second part of the trace of the full system run.}
    \label{fig:placeholder2}
\end{figure}

In Figures \ref{fig:placeholder1} and \ref{fig:placeholder2}, a full trace is provided, to illustrate the systems behavior. For clarity, some messages have been summarized or omitted, highlighting the key interactions and decision points while omitting routine or repetitive exchanges.

\end{document}